\documentclass{article}

     \usepackage[preprint, nonatbib]{neurips_2021}

\usepackage[utf8]{inputenc} %
\usepackage[T1]{fontenc}    %
\usepackage{hyperref}       %
\usepackage{url}            %
\usepackage{booktabs}       %
\usepackage{amsfonts}       %
\usepackage{nicefrac}       %
\usepackage{microtype}      %
\usepackage{xcolor}         %
\usepackage{amsmath}
\usepackage{amssymb}
\usepackage{algorithm}
\usepackage{algpseudocode}
\usepackage{graphicx}
\usepackage{multirow}
\usepackage{subcaption}
\usepackage{caption}
\usepackage{tcolorbox}
\usepackage{makecell}
\usepackage{todonotes}

\DeclareMathOperator*{\argmin}{\arg\,\min}
\newcommand{\norm}[1]{\left\Vert #1 \right\Vert}

\title{Distilled One-Shot Federated Learning}

\author{%
  Yanlin Zhou\thanks{Equal contribution.} \\
  NSF Center for Big Learning \\
  University of Florida \\
  \texttt{zhou.y@ufl.edu} \\
  \And
  George Pu$^*$ \\
  NSF Center for Big Learning \\
  University of Florida \\
  \texttt{pu.george@ufl.edu} \\
  \And
  Xiyao Ma\\
  NSF Center for Big Learning \\
  University of Florida \\
  \texttt{maxiy@ufl.edu} \\
  \AND
  Xiaolin Li\\
  Cognization Lab\\
  Palo Alto, CA\\
  \texttt{xiaolinli@ieee.org}\\
  \And
  Dapeng Wu \\
  NSF Center for Big Learning \\
  University of Florida \\
  \texttt{dpwu@ufl.edu} \\
}

\begin{document}

\maketitle

\begin{abstract}

Current federated learning algorithms take tens of communication rounds transmitting unwieldy model weights under ideal circumstances and hundreds when data is poorly distributed.
Inspired by recent work on dataset distillation and distributed one-shot learning, we propose Distilled One-Shot Federated Learning (DOSFL) to significantly reduce the communication cost while achieving comparable performance.
In just one round, each client distills their private dataset, sends the synthetic data (e.g. images or sentences) to the server, and collectively trains a global model.
The distilled data look like noise and are only useful to the specific model weights, \textit{i.e.,} become useless after the model updates.
With this weight-less and gradient-less design, the total communication cost of DOSFL is up to three orders of magnitude less than FedAvg while preserving between 93\% to 99\% performance of a centralized counterpart.
Afterwards, clients could switch to traditional methods such as FedAvg to finetune the last few percent to fit personalized local models with local datasets.
Through comprehensive experiments, we show the accuracy and communication performance of DOSFL on both vision and language tasks with different models including CNN, LSTM, Transformer, \textit{etc.}
We demonstrate that an eavesdropping attacker cannot properly train a good model using the leaked distilled data, without knowing the initial model weights. 
DOSFL serves as an inexpensive method to quickly converge on a performant pre-trained model with less than 0.1\% communication cost of traditional methods.

\end{abstract}

\section{Introduction}

Conventional supervised learning dictates that data be gathered into a central location where it can be used to train a model.
However, this is intrusive and difficult, if data is spread across multiple devices or clients.
For this reason, federated learning (FL) has garnered attention due to its ability to collectively train neural networks while keeping data private.
The most popular FL algorithm is FedAvg \cite{mcmahan2016communication}.
Each iteration, clients perform local training and forward the resulting model weights to a server.
The server averages these to obtain a global model.
Since learning processes happen at local level, neither the server nor other clients directly observe a client's data.

Federated learning introduces distinct challenges not present in classical distributed machine learning \cite{li2019federated}.
The main focus of this paper are expensive communication and statistical heterogeneity.
Previous approaches try to learn faster when data is poorly distributed \cite{abs/2103.11619}. 
They include modifying the training loss \cite{li2018federated}, using lifelong learning to prevent forgetting \cite{shoham2019overcoming}, and correcting local updates using control variates \cite{karimireddy2019scaffold}.
These methods improve upon FedAvg, but can still take hundreds of communication rounds, while increasing the amount of information sent to the server per round.

\begin{figure*}[t]
    \centering
    \includegraphics[width=\textwidth]{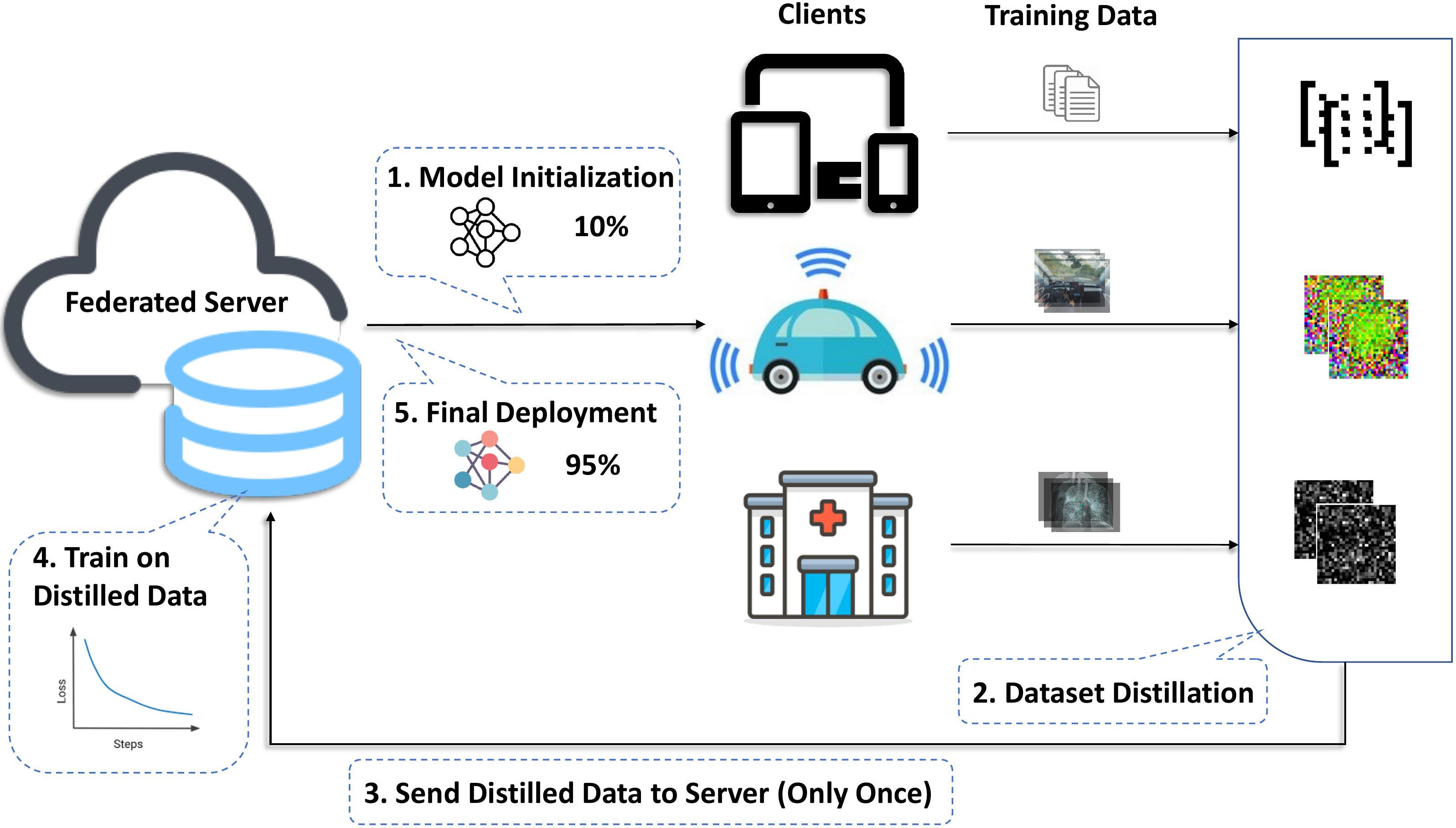}
    \caption{Distilled One-Shot Federated Learning. (1) The server initializes a model which is broadcast to all clients. (2) Each client distills their private dataset and (3) transmits synthetic data, labels and learning rates to the server. (4) The server fits its model on the distilled data and (5) distributes the final model to all clients.}
    \label{fig:dosfl}
\end{figure*}

Inspired by dataset distillation \cite{wang2018dataset}, we propose Distilled One-shot Federated Learning (DOSFL) (see Figure~\ref{fig:dosfl}) to reduce the communication cost by up to 3 orders of magnitude.
DOSFL requires only one round of communication between a server and its clients.
Each client distills their data and uploads learned synthetic data, label and learning rate to the server, instead of transmitting bulky gradients or weights.
Even large datasets containing thousands of examples can be compressed to only a few fabricated examples.
The server then interleaves the clients' distilled data together, using them to train a global model.
To achieve good results even when client data is poorly distributed, we leverage soft labels \cite{sucholutsky2019softlabel} and introduce two new techniques: soft reset and random masking.

Due to the reduction in rounds, we claim a communication reduction of up to 99.9\% compared to FedAvg while achieving similar accuracy.
DOSFL can also preserve between 93\% to 99\% of centralized training's performance when data is independently and identically distributed (IID).
Furthermore, under more challenging assumptions of low participation and asynchronous update with multiple rounds, DOSFL can achieve even better results and preserve higher centralized performance, \textit{i.e.,} almost 99\% on IID Federated MNIST.
Unlike dataset compression, fabricated inputs in dataset distillation (see Appendix \ref{app:images} and \ref{app:sents})---along with the corresponding labels and learning rate---are generated by a specific model weight distribution and become useless after the model updates.
We experimentally show that the final performance of a DOSFL is strongly dependent on the model parameters used to train on distilled data.
Without knowing the the exact initial weights distributed by the server, an eavesdropper cannot reproduce the global model with leaked distilled data.

We believe DOSFL is part of a new paradigm in the field of FL.
So far, nearly every FL algorithm communicates model weights or gradients.
While effective, breaking from this pattern offers many benefits, such as low communication \cite{guha2019one} or private model architectures \cite{li2019fedmd}.
We hope that DOSFL, along with related work, may inspire the machine learning community to explore possible techniques for weight-less and gradient-less FL.

\section{Related Work}

\subsection{Federated Learning}
Since the introduction of FedAvg in 2016 \cite{mcmahan2016communication}, there has been an explosion of work directed towards the problem of statistical heterogeneity.
When statistical hetergeniety is high, convergence for FedAvg slows and becomes unstable \cite{li2019federated}. 
The issue is that, the difference between the local losses and the global objective---their weighted sum---may be large.
As such, minimizing a particular local loss does not ensure that the global loss is also minimized.
This is problematic, even when the losses are convex and smooth \cite{li2019convergence}.
In applications where privacy loss can be tolerated, Zhao et al. demonstrate massive gains in performance 
by making as little as 5\% of data public \cite{zhao2018federated}.

Numerous successors to FedAvg have been suggested.
Server momentum introduces a global momentum parameter that improves convergence theoretically and experimentally \cite{liu2019accelerating}.
In Shoham et al., the loss is modified with Elastic Weight Consolidation to prevent forgetting as clients perform local training \cite{shoham2019overcoming}.
SCAFFOLD uses control variates, namely the gradient of the global model, to address drifting among clients during local training \cite{karimireddy2019scaffold}.
These schemes, while effective, at least double the per round communication cost. 
More efforts have not had this drawback.
The work in \cite{reddi2020adaptive} proposes several federated version of optimizers, FedAdaGrad, FedAdam and FedYogi, to provide better convergence for non-IID data.
The issue of objective inconsistency was discussed in \cite{wang2020tackling} and solved by the proposed FedNova, which averages the normalized local gradients.

While faster learning decreases the total number of communication rounds, strategies have been devised to explicitly reduce communication costs.
FedPAQ quantizes local updates before transmission, with averaging happens at both client and server sides \cite{reisizadeh2019fedpaq}.
Sparsifying the weights may perform better than FedAvg alone \cite{sattler2019robust}.
Asynchronous model updates have also been explored, using adaptive weighted averaging to combat staleness, combined with a proximal loss, \cite{xie2019asynchronous} and updating deeper layer less frequently than shallower layers \cite{chen2019communication}.
SAFA takes a semi-synchronous approach; only up-to-date and deprecated clients synchronize with the server \cite{wu2019safa}.

A few papers have made first steps towards one-shot FL. 
Guha et al. try different heuristics for selecting the client models which would form the best ensemble \cite{guha2019one}.
By swapping weight averaging for ensembling, only one round of communication is necessary.
Upper and lower bounds have been proven for one-shot distributed optimization, along with an order optimal algorithm \cite{salehkaleybar2019one, sharifnassab2019order}.
The extent to which these results apply to FL of neural networks is unknown as the local losses must be convex and drawn from the same probability distribution.
Recently, knowledge distillation has been introduced to the field of FL to reduce the computation costs for edge devices \cite{FedGKT2020}.
In addition, group knowledge transfer reduces communication costs by accelerating convergence.

\subsection{Distillation}

There is a wealth of literature studying dataset compression, while maintaining the most crucial features for training models.
These methods include dataset pruning \cite{angelova2005pruning} and core set construction \cite{bachem2017practical,tsang2005core,sener2017active}, which keep the examples that are measured to be more useful for training and remove the rest.
The drawback of drawing distilled images from the original dataset is that the level of compression achieved is much lower than that of dataset distillation, which is exempt from the requirement that distilled data be real \cite{wang2018dataset}.
Dataset distillation \cite{wang2018dataset} was introduced by Wang et al., to compress a large dataset with thousands to millions of images down to only a few synthetic training images.
The key idea is to use gradient descent to learn the features most helpful for rapidly training a neural network.
Given some model parameters $\theta_0$, dataset distillation minimizes the loss of adapted parameters $\theta_1$, obtained by performing gradient descent on $\theta_0$ and the distilled data.
This procedure resembles meta-learning, which performs task-specific adaption followed by a meta-update \cite{finn2017model}.
With dataset distillation, 10 synthetic digits can train a neural network from 13\% to 94\% test accuracy in 3 iterations, near the $98\%$ test accuracy reached by training on MNIST.

Dataset distillation originally was limited to only image classification tasks, because the distilled labels were predetermined and fixed.
Learnable or soft labels not only decrease the number of required labels, but also expand dataset distillation to language tasks such as sentiment classification \cite{sucholutsky2019softlabel}.
Soft labels have a long history, being proposed for model distillation by Hinton et al. \cite{hinton2015distilling} and for k-nearest neighbors by El Gayar et al. \cite{gayar2006study}.
Using soft label dataset distillation, Sucholutsky et al. were able to train LeNet to $96\%$ accuracy with only 10 images \cite{sucholutsky2019softlabel}.
Examples of distilled data (i.e., text, grey image and RGB image) are shown in Figure~\ref{fig:dosfl}.

\section{Distilled One Shot Federated Learning}

Suppose we have numbered clients $1, \dots, N$ each with their own local models $f_{\theta_1}, \dots, f_{\theta_N}$ with parameters $\theta_1, \dots, \theta_N$ and loss functions $L_1, \dots, L_N$.
Given some probability vector $p = (p_1, \dots, p_N)$ (each $0 \leq p_k \leq 1$ and $\sum_k p_k = 1$), our goal is to find some parameters $\theta^*$ that minimize the weighted sum $L = \sum_{k=1}^{N} p_k L_k(\theta)$.
\begin{equation}
    \theta^* = \argmin_{\theta} \sum_{k=1}^{N} p_k L_k(\theta)
\end{equation}

However, we often do not have distinct loss functions but rather the same loss function $\ell$ evaluated on distinct private datasets.
Let $\ell(\theta; x, y)$ to be the loss of a single example $(x, y)$.
Following \cite{wang2018dataset}, we define $\ell(\theta; \{(x_i, y_i)\})$ to be average loss of all the data points in the set $\{(x_i, y_i)\}$.
Thus, for each client $k = 1, \dots, N$ with a dataset $\mathcal{X}_k$, $L_k(\theta) = \ell(\theta; \mathcal{X}_k)$.

\begin{algorithm}[t]
    \caption{Distilled One-Shot Federated Learning}
    \label{alg:dosfl}
    \begin{algorithmic}[1]
        \State Initialize server weights $\theta_0$
        \For{clients $k=1, \dots, N$}
            \State \Call{DistillData}{$k$, $\theta_0$}
            \State Send distilled data to the server
        \EndFor
        \State Merge clients' distilled data into a single sequence $\{(\tilde{x}_j, \tilde{y}_j, \tilde{\eta}_j)\}_{j=1}^{NS_d}$
        \For{$i = 0, 1, \dots, E_d - 1$}
            \For{$j = 0, 1, \dots, NS_d$}
                \State Number of adaptations $a = iNS_d + j$
                \State $\theta_{a+1} \gets \theta_a - \tilde{\eta}_j \nabla \ell(\theta_a; \tilde{x}_j, \tilde{y}_j)$
            \EndFor
        \EndFor
        \State
        \Function{DistillData}{$k$, $\theta_0$}
            \State Initialize $\{(\tilde{x}_j, \tilde{y}_j, \tilde{\eta}_j)\}_{j=1}^{S_d}$
            \For{$e = 1, 2, \dots, E$}
                \State Get a minibatch $\mathcal{B}$ from client's dataset $\mathcal{X}_k$
                \For{$i = 0, 1, \dots, E_d - 1$}
                    \For{$j = 0, 1, \dots, S_d - 1$}
                        \State Number of adaptations $a = iS_d + j$
                        \State $\theta_{a+1} = \theta_a - \tilde{\eta}_j \nabla \ell(\theta_a; \tilde{x}_j, \tilde{y}_j)$
                    \EndFor
                \EndFor
                \State $\tilde{x} \gets \tilde{x} - \alpha \nabla_{\tilde{x}} \, \ell(\theta_{E_dS_d}; \mathcal{B})$
                \State $\tilde{\eta} \gets \tilde{\eta} - \alpha \nabla_{\tilde{\eta}} \, \ell(\theta_{E_dS_d}; \mathcal{B})$
                \If{using soft labels}
                    \State $\tilde{y} \gets \tilde{y} - \alpha \nabla_{\tilde{y}} \, \ell(\theta_{E_dS_d}; \mathcal{B})$
                \EndIf
            \EndFor
            \State \Return $\{(\tilde{x}_i, \tilde{y}_i, \tilde{\eta}_i)\}_{i=1}^{S_d}$
        \EndFunction
    \end{algorithmic}
\end{algorithm}

Our solution consists of 3 steps. 
These steps are summarized in Algorithm~\ref{alg:dosfl}.
\begin{enumerate}
    \item A central server randomly initializes model parameters $\theta_0$.
    This can be distributed to the clients as a random seed.
    \item The clients distill their datasets.
    Start by initializing the distilled data $\tilde{x}$, distilled label $\tilde{y}$, and distilled learning rate $\tilde{\eta}$.
    Each entry in $\tilde{x}$ is drawn from a standard normal distribution, while $\tilde{\eta}$ is set to a predefined value $\eta_0$.
    The distilled labels are initialized as either one-hot vectors for classification problems or normal distributed random vectors for regression problems.
    Adapt these into $\theta_1$ via gradient descent.
    \begin{equation}
        \theta_1 = \theta_0 - \tilde{\eta} \ell(\theta_0; \tilde{x}, \tilde{y})
    \end{equation}
    Afterwards, minimize the loss of $\theta_1$ evaluated on a batch of real data $\mathcal{B}$.
    
    This can be done with a sequence of distilled data $\{(\tilde{x}_j, \tilde{y}_j, \tilde{\eta}_j)\}_{j=1}^{S_d}$ ($S_d$ is the distill step), repeated distill epoch $E_d$ times.
    Each example successively adapts $\theta_0$ until we have $\theta_{E_dS_d}$ after $E_dS_d$ gradient descent updates.
    This dramatically increases the expressive power of dataset distillation at the expense of compute time.
    \item The clients upload the distilled data to the server.
    If $S_d > 1$, the server sorts the distilled data by index, e.g. $\{x_1, x_2, x_3\}, \{y_1, y_2, y_3\}$ from clients 1 and 2 become $\{x_1, y_1, x_2, y_2, x_3, y_3\}$ where $x_j, y_j$ are 3-tuples.
    The server then trains its own model on the combined sequence.
\end{enumerate}

The last step can cause issues when the data is non-IID.
Consider two clients $1$ and $2$ with distilled examples $x_1$ and $y_1$ respectively with $E_d = 1$.
The server first trains $\theta_0$ on $x_1$, arriving at $\theta_1$, which is then trained on $y_1$.
But $y_1$ has been distilled to train $\theta_0$.
To combat this interference, we introduce two new techniques for improving performance on non-IID data.

Soft resets sample the starting parameters, $\theta_0$ from a Gaussian distribution, around the server's parameters $\mathcal{N}(\theta_0, \sigma_{sr}^2)$.
By sampling $\theta_0$ between distillation iterations, dataset distillation learns more robust examples capable of training any model with weights $\theta \sim \mathcal{N}(\theta_0, \sigma_{sr}^2)$.
This technique is based off of the ``hard resets'' introduced in \cite{wang2018dataset}, which completely re-initializes $\theta_0$.
Data distilled with ``hard resets'' can be used on any randomly initialized model, but cannot train models to the same level of accuracy as models trained on data distilled without resets.

Random masking randomly selects a fraction $p_{rm}$ of the distilled data at each training iteration and replaces it with a random tensor.
The random tensors randomly adjusts the model during training, while also reducing the amount of distilled data to actually train the starting parameters.
After the training iteration, the original distilled data are restored.
Now, sequences of distilled data can still train a model even when there is interference from other distilled steps.
However, resetting and storing distilled data is compute and memory intensive, which slows down distillation.

\section{Experiments}\label{sec:exp}

We evaluate DOSFL on several federated classification tasks.
Because of this, cross entropy loss is used for all experiments.
To train the distilled data, we use ADAM \cite{kingma2014adam} with a learning rate $\alpha$ that is halved every $\tau$ epochs.
We have $\alpha = 0.01, \tau = 40$, $\alpha = 0.01, \tau = 10$, and $\alpha = 0.1, \tau = 30$ for federated MNIST, IMDB, and TREC-6 respectively.
These hyperparameters are mirrored from \cite{sucholutsky2019softlabel} and have been found to be near-optimal.
For federated Sent140, we replicated the hyperparmeters from federated IMDB. 
Clients distill the data for $E = 30$ epochs for image datasets and $E = 50$ epochs for text datasets with a batch size of $B = 512$.
All experiments were run on an Nvidia M40 GPU and Intel Xeon E5-2695 CPU, taking 2-3 minutes per client for 100 client federated MNIST, $<1$ minute per client for 29 client federated IMDB and 100 client TREC-6, and 14-15 minutes for 100 client federated Sent140.
We use the default train and test splits associated with each dataset.

The client-server architecture is simulated by partitioning a dataset into subsets, and then distilling these subsets.
The server models have their weights Xavier initialized \cite{glorot2010understanding}.
The weights are then replicated across each client.
Following the methodology of McMahan et al., IID partitions are created by randomly dividing the dataset into subsets \cite{mcmahan2016communication}.
For non-IID partitions, we first sort the entire dataset by label and then divide it into $Ns$ shards of equal length.
Starting from $N$ empty subsets, the shards are randomly assigned to the subsets until each has $s$ shards.
As $s$ increases, the partition becomes more IID, with subsets more likely to contain examples from each class.

\subsection{Image Classification}\label{sec:exp_image}

\begin{table*}[t]
    \centering
    \caption{DOSFL accuracy on federated MNIST. For reference, LeNet can reach a test set accuracy of $>99\%$ on MNIST and 93\% with dataset distillation. The highest value in each column is \textbf{bold}.}
    \label{tab:ablation_mnist}
    \begin{tabular}{r c c c c}
        \toprule
        Additions & \multicolumn{2}{c}{10 clients} & \multicolumn{2}{c}{100 clients} \\
        \cmidrule(l){2-3} \cmidrule(l){4-5}
                        & IID & non-IID & IID & non-IID \\
        \midrule
        None            & 94.02\% & $51.55\%$ & $88.27\%$ & $59.25\%$\\
        Soft label (SL) & \textbf{95.64\%} & $61.26\%$ & $\textbf{91.53\%}$ & $63.32\%$\\
        Soft reset (SR) & 93.05\% & $77.71\%$ & $88.41\%$ & $\textbf{78.29\%}$\\
        Random masking (RM) & 93.68\% & $70.13\%$ & $87.34\%$ & $ 69.87\%$\\
        SL, SR          & 90.87\% & $\textbf{78.83\%}$ & $85.54\%$ & $76.55\%$ \\
        SR, RM          & 88.07\% & $78.54\%$ & $82.38\%$ &  $73.62\%$\\
        SR, RM, SL      & 88.95\% & $78.22\%$ & $83.71\%$ & $77.58\%$  \\
        \bottomrule
    \end{tabular}
\end{table*}

We first test DOSFL on 10 and 100 client Federated MNIST with different combinations of soft labels, soft resets, and random masking.
All federated MNIST experiments use LeNet as the model architecture \cite{lecun1998gradient}.
Our distilled data are not single examples but batches with size $B_d = 10$.
Within each batch, the labels are initialized to one of each class, \textit{e.g.} one label is `1', one label is `2', etc.
Using soft labels, $B_d$ could be made much smaller without loss of performance.
After the server model has trained on distilled data from the clients, its accuracy is measured on a test set.
Each experiment is run $5$ times, and the best result is reported in Table~\ref{tab:ablation_mnist}.

The distill steps are $S_d = 30$, the distill epochs are $E_d = 3$, and the initial distill learning rate is $\eta_0 = 0.02$.
Of the proposed additions to dataset distillation, soft resets provide the largest jump in non-IID performance, followed by random masking and soft labels.
The reset variance $\sigma^2_{sr} = 0.2$ and masking probability $p_{rm} = 0.3$ were chosen from the search spaces $\{0.1, 0.2, \dots, 0.6\}$ and $\{0.1, 0.2, 0.3, 0.5, 1\}$.
However, soft labels also boost accuracy when data is IID, where as the other two methods cause dips in the final accuracy.
The distillation additions are not additive; even with all add-ons, non-IID DOSFL caps at $\sim 79\%$ test accuracy.
Surprisingly, the behavior of these additions changes depending on the number of clients.
While accuracies in the 100 client case are lower in general, soft resets and no additions work better.

\begin{table*}[t]
    \centering
    \caption{DOSFL performance on all federated learning tasks. Instead of 10 and 100 clients for \textbf{TREC-6}, we have \textbf{2} and \textbf{29} respectively. For the comparison with non-IID FedAvg, we \textbf{bold} the greater accuracy.}
    \resizebox{\textwidth}{!}{
    \label{tab:imdb+trec}
    \begin{tabular}{l c c c c c c c c}
        \toprule
        Dataset & \multicolumn{3}{c}{10 clients} & \multicolumn{3}{c}{100 clients} &
        \multicolumn{2}{c}{Centralized} \\
        \cmidrule(l){2-4} \cmidrule(l){5-7} \cmidrule(l){8-9}
        & IID & \multicolumn{2}{c}{non-IID} & IID & \multicolumn{2}{c}{non-IID} \\
        \cmidrule(l){3-4} \cmidrule(l){6-7}
        & DOSFL & DOSFL & FedAvg & DOSFL & DOSFL & FedAvg & DOSFL & - \\
        \midrule
        MNIST & $90.87\%$ & \textbf{78.83\%} & $44.01\%$ & $85.54\%$ & $76.55\%$ & \textbf{95.26\%} & $93.61\%$ & $98.86\%$ \\
        IMDB & $81.04\%$ & \textbf{79.86\%} & $50.02\%$ & $71.94\%$ & \textbf{70.75\%} & $60.45\%$ & $78.3\%$ & $86.1\%$ \\
        TREC-6 & $83.60\%$ & \textbf{73.40\%} & $49.80\%$ & $79.00\%$ & \textbf{73.60\%} & $13.70\%$ & $79.2\%$ & $89.4\%$ \\
        Sent140 & $78.10\%$ & \textbf{78.00\%} & $36.55\%$ & $73.50\%$ & \textbf{69.50\%} & $52.58\%$ & $76.62\%$  & $80.1\%$ \\
        \bottomrule
    \end{tabular}
    }
\end{table*}

\subsection{Text Classification}

To show that DOSFL is not limited to image-based tasks, we test DOSFL on federated IMDB (sentiment analysis) \cite{maas-EtAl:2011:ACL-HLT2011}, federated TREC-6 (question classification) \cite{voorhees2000overview}, and federated Sent140 (sentiment analysis) \cite{go2009twitter}.
Directly applying dataset distillation for language tasks is challenging as text data is discrete.
Each token is a one-hot vector with dimension equal to the number of tokens, which can be in the thousands.
To overcome this issue, we use pre-trained GloVe embeddings with a look up table to convert one-hot token ids to word vectors in 100D Euclidean space \cite{pennington2014glove}.
Distilled sentences now are fixed-size real-valued matrices.
Real sentences are also padded or truncated to the same fixed length: 200 for federated IMDB and Sent140, 30 for federated TREC-6.

The results, best out of 5 runs, are provided in Table~\ref{tab:imdb+trec}.
We also provide the accuracy of non-IID FedAvg after an equivalent amount of communication.
This allows DOSFL and FedAvg to be compared in an equal communication cost setting.
We tuned the FedAvg hyperparameters to maximize initial learning: local epochs $E = 5$, batch size $B = 10$, and learning rate $\eta = 0.01$.
The batch size was the lowest in the considered range $\{10, 20, 50, 100, 200, 500\}$ and the learning rate the largest in $\{10^{-2}, 3 \times 10^{-3}, 10^{-3}, 3 \times 10^{-4}, 10^{-4}\}$.
Note that, the preserved accuracy, defined as the DOSFL accuracy over baseline performance, of all tasks is at least $93\%$.
This shows that DOSFL is capable of handling different tasks with small or large datasets in one round.

For federated IMDB, we use a simple CNN model called TextCNN.
We test DOSFL with soft labels for 10 and 100 clients, IID and non-IID federated IMDB.
Here the distill steps $S_d = 5$, the distill epochs $E_d = 10$, the distill batch size $B_d = 1$, and the starting distill learning rate is $\eta_0 = 0.01$.
Since there exists only 2 classes in IMDB dataset (positive or negative sentiment), non-IID performance is within 2\% of IID.
Approximately $\nicefrac{3}{4}$ clients contain labels from all classes, whereas in federated MNIST no client can have more than 4 classes.

For federated TREC-6, we adopt a Bi-LSTM model to show that DOSFL can be used with non-CNN models.
We use 2 and 29 for the number of clients, since the size of the dataset is 5452 and the client dataset sizes must be divisible by the shard count $s = 2$.
The amount of training data for the 2 client federated TREC-6 and 10 client federated IMDB are almost equal ($2726 \sim 2500$).
Similarly, 29 client federated TREC-6 is comparable with 100 client federated IMDB ($188 \sim 250$).
We have $S_d = 2$, $E_d = 1$, $B_d = 1$, and $\eta_0 = 1.5$.
Due to the low number of clients, we were able to reduce the amount of distilled data needed compared to the previous two tasks.
Unlike federated IMDB, there is a larger $\sim 6\%$ gap in accuracy between the IID and non-IID settings.
Furthermore, we extend DOSFL to a larger dataset, Sent140, using TextCNN.
The hyperparameters are distill steps $S_d = 5$, distill epochs $E_d = 15$, $B_d = 1$, and initial learning rate $\eta_0 = 0.3$.

\subsection{DOSFL with Stragglers and Low Participation}\label{sec:low_parti}

\begin{figure}[t]
    \centering
    \begin{subfigure}{0.48\linewidth}
        \centering
        \includegraphics[width=\linewidth]{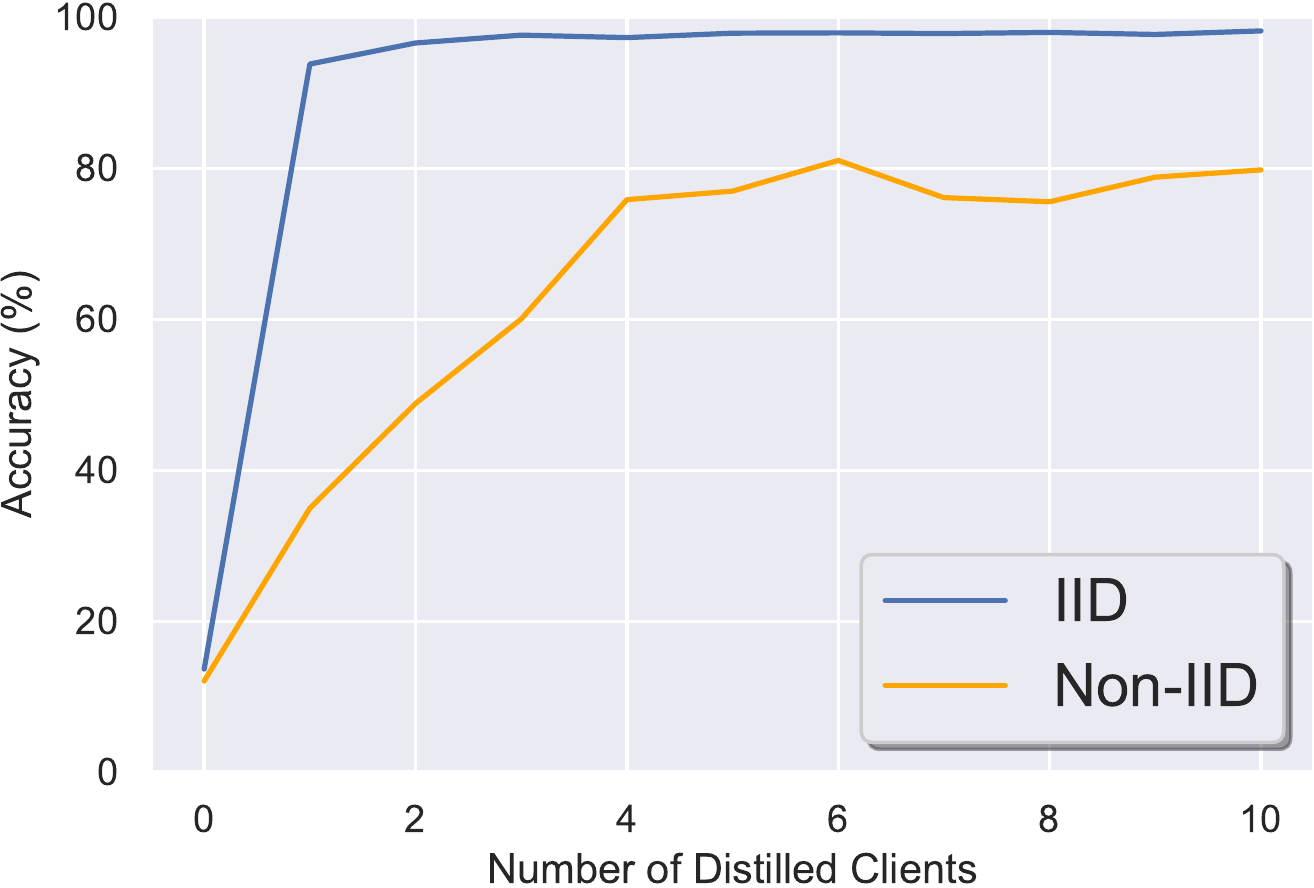}
        \caption{Accuracy for 10 clients. The final accuracies \protect\newline are $98.85\%$ and $79.79\%$.}
    \end{subfigure}
    \begin{subfigure}{0.48\linewidth}
        \includegraphics[width=\linewidth]{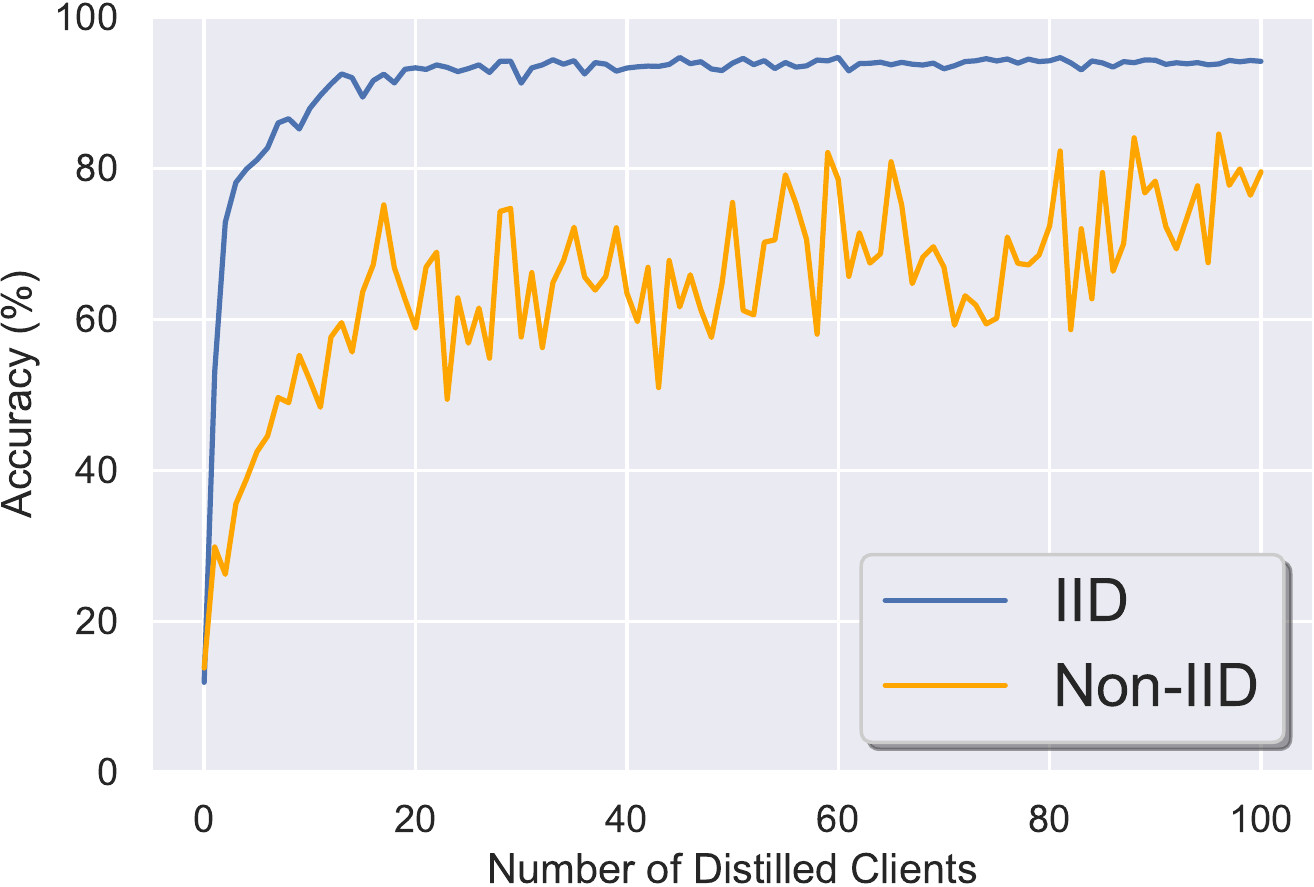}
        \caption{Accuracy for 100 clients. The final accuracies \protect\newline are $94.28\%$ and $79.53\%$.}
    \end{subfigure}
    \caption{Performance of LP-DOSFL with low participation on Federated MNIST, with soft resets and soft labels, vs. the number of clients distilled.} 
    \label{fig:acc_clients}
\end{figure}

So far The above mentioned settings assumes synchronous full participation which is highly unlikely in the real-world.
We design an alternate version of DOSFL for when the participation rate is extremely low such that client communication is almost one-by-one.
First, the server selects only one client to distill its data.
Afterwards, the server updates the global model by training on the distilled data.
A different client then performs dataset distillation targeting the updated parameters.
This process repeats until each client has distilled their data.
Hence, the global model is updated $N$ times, once for each client.
The communication is highly serial; the next client can only begin distillation after the current client finishes.

We name this setting LP-DOSFL and evaluate its performance on MNIST in Figure~\ref{fig:acc_clients}, using the best out of 5 trials again.
Surprisingly, LP-DOSFL achieves almost 99\% accuracy when MNIST is IID, $7\%$ more than vanilla DOSFL.
This advantage disappears when the data becomes non-IID.
The final accuracy of LP-DOSFL is only $1\%$ larger than plain DOSFL.
The reason for this is simple: when the clients' datasets are IID, a model trained on one client's dataset will transfer to the others.
Encouragingly, the server model achieves its final accuracy after as few as 15\% of the clients finish dataset distillation.
This is true even when the data is non-IID, although it takes longer (around 40\% of the clients).
Thus, the total amount of communication needed for LP-DOSFL is less than DOSFL.

\section{Discussion}\label{sec:disc}

\subsection{Communication}

\begin{table*}[t]
    \centering
    \caption{Communication comparison between DOSFL and FedAvg. The accuracy at the break even round is given for FedAvg on the non-IID partition among 100 clients.}
    \label{tab:comcost}
    \begin{tabular}{l l c c c}
        \toprule
        Dataset & Model & Data size & Model size & Break even round \\
        \midrule
        MNIST & LeNet & $28 \times 28$  & $61,706$ & $19.83$ \\
        IMDB & TextCNN & $200 \times 100$ & $120,601$ & $8.29$ \\
        TREC-6 & Bi-LSTM & $30 \times 100$ & $404,406$ & $0.14$ \\
        Sent140 & TextCNN & $200 \times 100$ & $120,601$ & $8.29$ \\
        Sent140 & Transformer & $200 \times 100$ & $13,151,238$ & $0.54 $ \\
        \bottomrule
    \end{tabular}
\end{table*}

\begin{table*}[t]
    \centering
    \caption{Communication comparison between DOSFL and FedAvg. The accuracy at the break even round is given for FedAvg on the non-IID partition among 100 clients.}
    \label{tab:comcost2}
    \begin{tabular}{l l c c c c}
        \toprule
        Dataset & Model & \multicolumn{2}{c}{10 Clients} & \multicolumn{2}{c}{100 Clients} \\
        \cmidrule(l){3-4} \cmidrule(l){5-6}
        & & \thead{Comm. \\ Reduction} & \thead{DOSFL to \\ FedAvg Ratio} & \thead{Comm. \\ Reduction} & \thead{DOSFL to \\ FedAvg Ratio}\\
        \midrule
        MNIST & LeNet & \multicolumn{2}{c}{FedAvg does not converge} & N/A & 0.44 \\ %
        IMDB & TextCNN & \multicolumn{2}{c}{FedAvg does not converge} &  $66.83\%$ & 3.0 \\ %
        TREC-6 & Bi-LSTM & $99.22\%$ & 127.76 &  $99.71\%$ & 356.4 \\ %
        Sent140 & TextCNN & $87.98\%$ & 8.3 & $93.26\%$ & 14.8\\ %
        Sent140 & Transformer & \multicolumn{2}{c}{FedAvg does not converge} & $ 99.90\%$ & 1012.5 \\ %
        \bottomrule
    \end{tabular}
\end{table*}

We now compare the total communication cost (TCC) of DOSFL with that of FedAvg measured in the amount of scalar values sent between the clients and the server.
Since the server model's initialization can be distributed as a random seed, we ignore the cost of the first server-to-client transmission.
Let $C = 0.1$ be the fraction of the $N$ clients that participate each round.
FedAvg sends $\Theta$ server model parameters to each client, who responds with locally trained parameters.
Let $T$ be the number of communication rounds.
\begin{equation}
    TCC_{fedavg} = NC\Theta (2T - 1)
\end{equation}
For DOSFL, we only need to consider the single expense of sending distilled data to the server.
Let $n_{data}$ be the number of elements in each data point and $B_d$ be the batch size of the distilled data.
\begin{equation}
    TCC_{dosfl} = NS_d(n_{data} + n_{cls} + 1)B_d
\end{equation}

Both formulas can be used to calculate the number of communication rounds---the break even round---needed for lifetime cost of FedAvg to equal DOSFL for the tasks in Section~\ref{sec:exp}.
\begin{equation}
    T_{break\ even} = \frac{1}{2} \left( \frac{S_d(n_{data} + n_{cls} + 1)B_d}{C\Theta} + 1 \right)
\end{equation}
Note that this value is independent of the number of clients $N$.
Break even rounds for federated MNIST, IMDB, TREC-6, and Sent140 are provided in Table~\ref{tab:comcost} along with the data and model size.
We also investigate potential communication savings when using larger models, such as Transformers, for Sent140. 
The higher break even round for MNIST, compared to the text tasks, is due to LeNet having significantly fewer parameters than either TextCNN or Bi-LSTM.
The best accuracy at the break even round $T_{break\ even}$ is reported in Table~\ref{tab:imdb+trec}.

Finally, we conclude our discussion of communication efficiency by comparing DOSFL with FedAvg under an iso-accuracy setting.
Let $T_{iso\ accuracy}$ be the number of communication rounds required for FedAvg to reach the accuracy achieved in Table~\ref{tab:imdb+trec}.
Define the DOSFL to FedAvg ratio as
\begin{equation}
    ratio = \frac{TCC_{dosfl}}{TCC_{fedavg}} = \frac{S_d(n_{data} + n_{cls} + 1)B_d}{NC\Theta (2T_{iso\ accuracy} - 1)}
\end{equation}
We can also calculate the percent communication reduction using the above ratio.
\begin{equation}
    reduction = \frac{TCC_{fedavg} - TCC_{dosfl}}{TCC_{fedavg}} = 1 - \frac{1}{ratio}
\end{equation}
We choose the smallest iso-accuracy round $T_{iso\ accuracy}$ out of 5 trials.
The values for both quantities is shown in Table~\ref{tab:comcost2}.
For some tasks, FedAvg fails to converge due to having only 1 client update the gradient each round.
In general, DOSFL saves more communication when the size of the model increases or when the dataset becomes more challenging.
Besides MNIST, communication savings are up to 3 orders of magnitude.

DOSFL provides an efficient way to trade computation and a few-to-none percentage points of accuracy for a great amount of communication cost reduction.
Next, clients can choose to either adopt another Federated Learning algorithm to continue improving the global model or personalize by training on local data.
As such, DOSFL is best suited for cross-silo FL, where 2-100 organizations seek to learn a shared model without sharing data \cite{kairouz2019advances}.
In cross-silo learning, participants likely would be able to dedicate hardware for the sole purpose of FL. 
Big models are also probable, and the communication savings of DOSFL increase as the models get larger.
Computation resources are also cheaper to acquire compared to communication resources.
A company can purchase dedicated GPUs and use them for years at reasonable cost without losing much of their intrinsic value.

\subsection{Privacy and security}

\begin{figure}
    \centering
    \begin{subfigure}[]{0.49\linewidth}
    \centering
        \includegraphics[width=\linewidth]{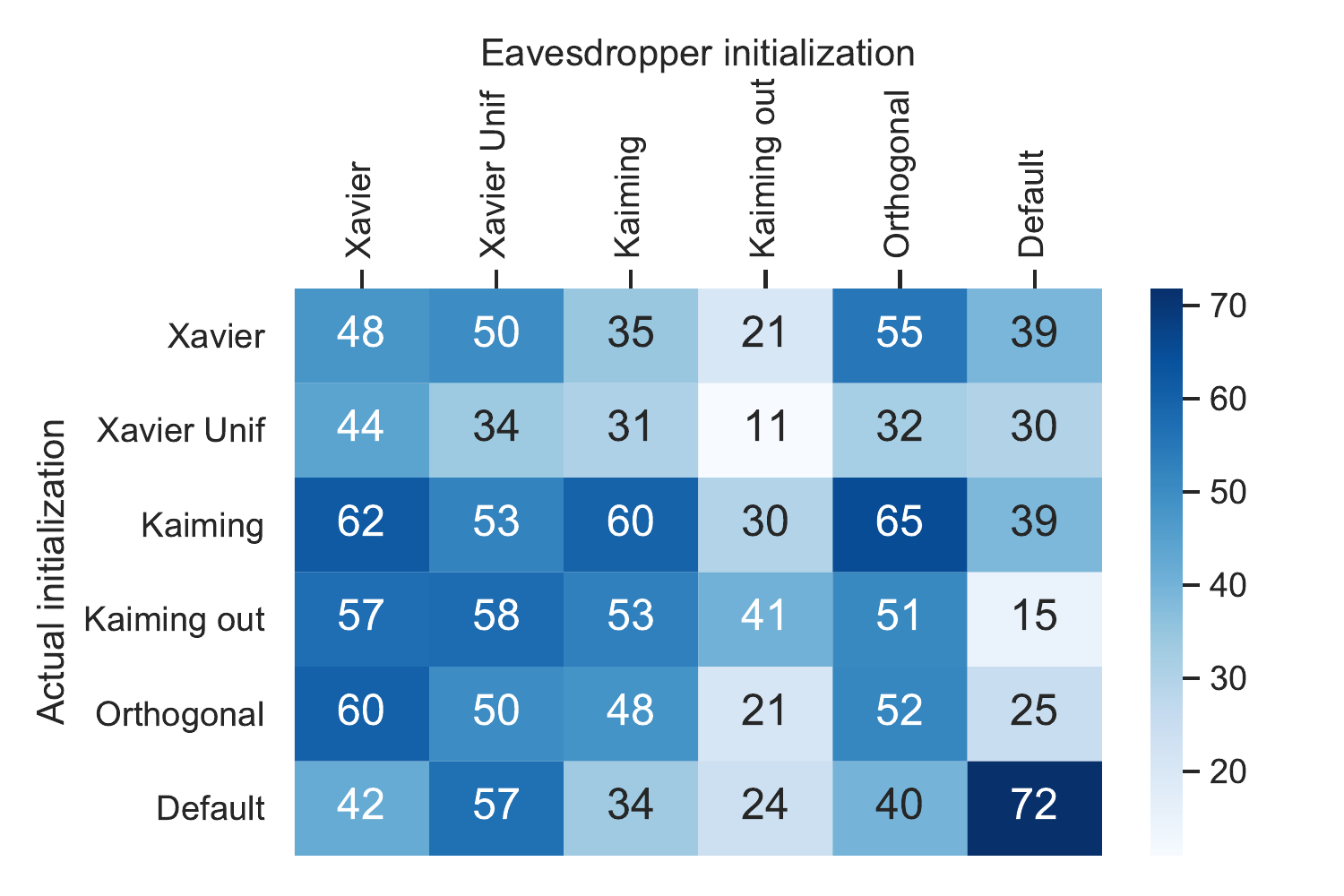}
        \caption{Performance matrix of eavesdroppers training on leaked distilled data without knowing the initialization method used for distillation. }
        \label{fig:attack}
    \end{subfigure}
    \begin{subfigure}[]{0.49\linewidth}
        \centering
        \includegraphics[width=\linewidth]{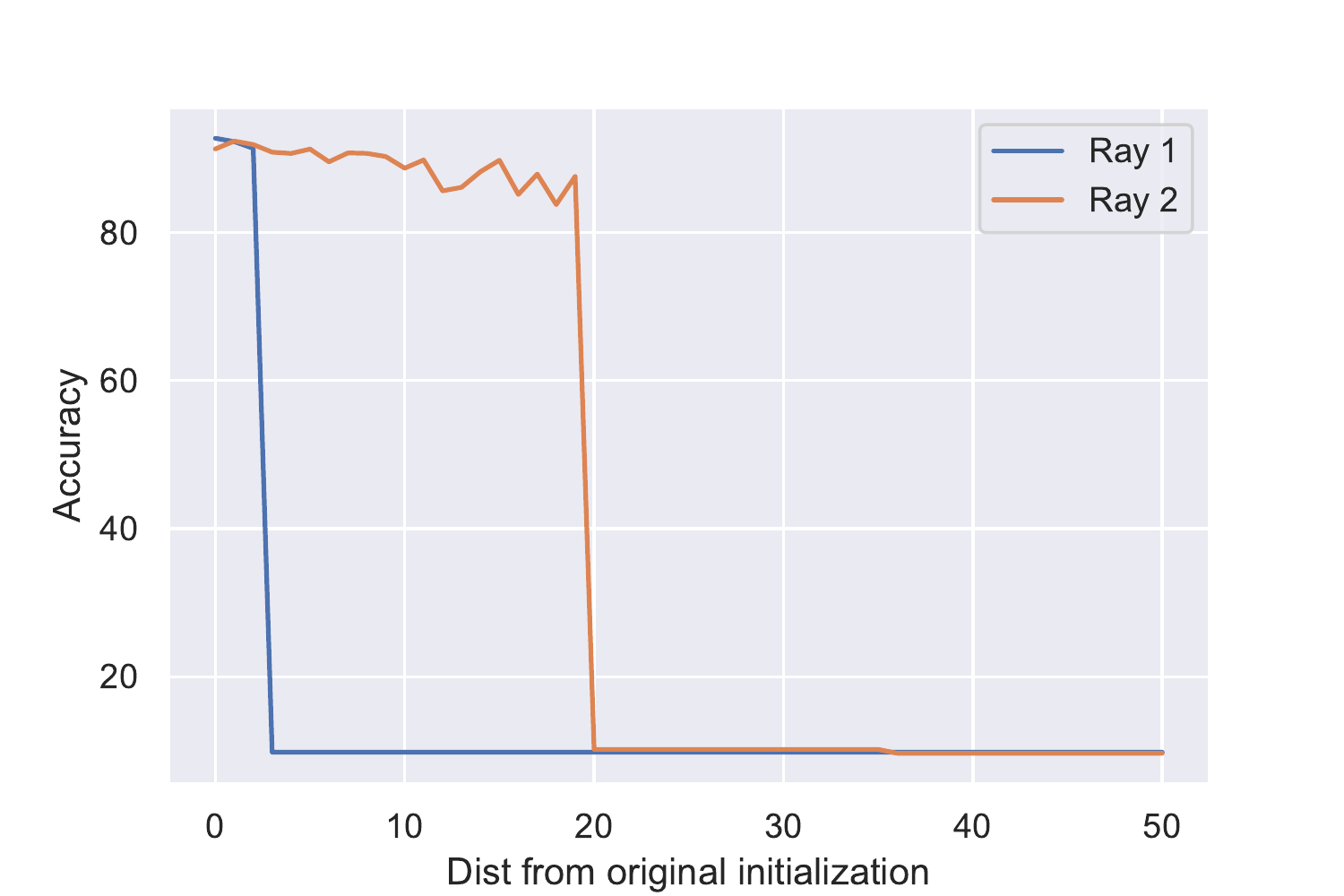}
        \caption{Accuracy of eavesdropper on leaked distilled data vs. distance from initialized weights in two random directions.}
        \label{fig:attack2}
    \end{subfigure}
    \caption{Privacy and security analysis of DOSFL.}
\end{figure}

Suppose the server or a client attempted to learn private information about other clients.
To the human eye, the distilled images or text appear random (see Appendix~\ref{app:images}, \ref{app:sents}).
So the only known way to extract information from the distilled data is train the targeted model.
Because distilled data are targeted towards a \emph{specific} initialization, training a differently initialized model on distilled data should result in a low accuracy model.
Therefore, we examine the security of DOSFL against an eavesdropping attack and show that an attacker cannot reproduce the global model without the server's initial weights.
Referring to Figure~\ref{fig:dosfl}, assume that the attacker intercepts the distilled data, labels and learning rates from each client between Step 2 and 3.

We assume that the model initialization, Step 1, is unknown to the attacker.
This can be easily achieved by being predefined offline or via encryption.
In Figure~\ref{fig:attack}, the attacker does not have access to the model distribution and tries to guess different initialization methods.
All results are averages of 5 trials.
These include Xavier, Xavier uniform, Kaiming, Kaiming uniform, orthogonal and the default PyTorch initialization methods \cite{NEURIPS2019_9015}.
All attacks fail to reproduce the same level of performance of the global model as most attacks struggle to reach over 50\% accuracy.
The accuracy of the eavesdropper drops off with increased distance from the original model.
Given the initial weights $\theta_0$ and a vector $v$ whose components follow a standard Normal distribution, we can create perturbed parameters $\tilde{\theta}_0 = \theta_0 + k v/\norm{v}$ that is $k$ distance away.
As seen in Figure~\ref{fig:attack2}, the final accuracy decreases suddenly with increased distance from the original weights.

As such, the privacy of DOSFL should be \emph{no worse} than plain FedAvg.
There are some security risks from other types of attacks.
In \cite{wang2018dataset}, the authors used dataset distillation for data poisoning attacks.
Images were distilled such that, after one gradient descent step, the final model would misclassify the attack category.
However, this security risk is also present with FedAvg and most other FL algorithms: clients can deliberately upload poisoned weights to the server \cite{bagdasaryan2020backdoor}.
For DOSFL, bounding the gradient value or using momentum when training the global model on distilled data could mitigate this threat.
While defenses against other types of attacks are beyond the scope of this paper, we believe that existing differential privacy and secure multi-party computation tools will prove sufficient.

\newpage
\bibliographystyle{plain}
\bibliography{refs}

\newpage
\appendix

\section{Hyperparameters}

\begin{table}[h]
    \centering
    \caption{DOSFL hyperparameters. A dash (--) indicates that the value is constant across different tasks.}
    \label{tab:hyperparams}
    \begin{tabular}{l c c c c c}
        \toprule
        Hyperparameter & Symbol & \multicolumn{4}{c}{Value} \\
        \cmidrule(l){3-6} & & MNIST & IMDB & TREC6 & Sent140\\
        \midrule
        Batch size & & 512 & -- & -- & -- \\
        Distill batch size & $B_d$ & $40$ & $1$ & $1$ & $1$\\
        Distill steps & $S_d$ & $30$ & $5$ & $2$ & $5$  \\
        Distill epochs & $E_d$ & $3$ & $10$ & $1$ & $15$ \\
        Initial distilled learning rate & $\eta_0$ & $0.02$ & $0.01$ & $1.5$ & $0.3$ \\
        Learning rate & $\alpha$ & $0.01$ & $0.01$ & $0.1$ & $0.01$ \\
        Learning rate decay & & $0.5$ & -- & -- & -- \\
        Learning rate decay period & $\tau$ & $40$ & $10$ & $30$ & $10$ \\
        Local epochs & $E$ & $30$ & $50$ & $50$ & $50$ \\
        Random masking probability & $p_{rm}$ & $0.3$ & -- & -- & -- \\
        Soft reset variance & $\sigma_{sr}^2$ & $0.2$ & -- & -- & -- \\
        \bottomrule
    \end{tabular}
\end{table}

\section{Distilled Data}

\subsection{MNIST}\label{app:images}

\begin{figure}[h]
    \centering
    \includegraphics[width=\linewidth]{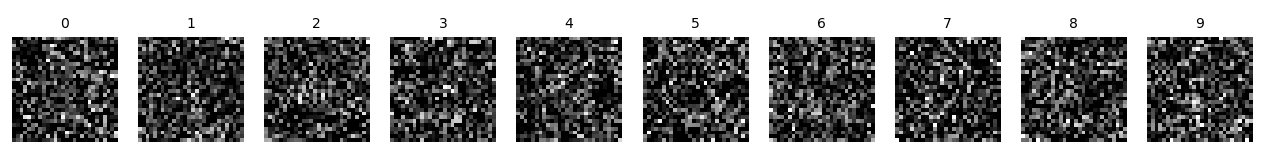}
    \caption{First step of distilled images from 1 out of 10 clients for IID federated MNIST with no additions (i.e. soft labels, soft resets, random masking).}
    \label{fig:hard_iid}
\end{figure}

\begin{figure}[h]
    \centering
    \includegraphics[width=\linewidth]{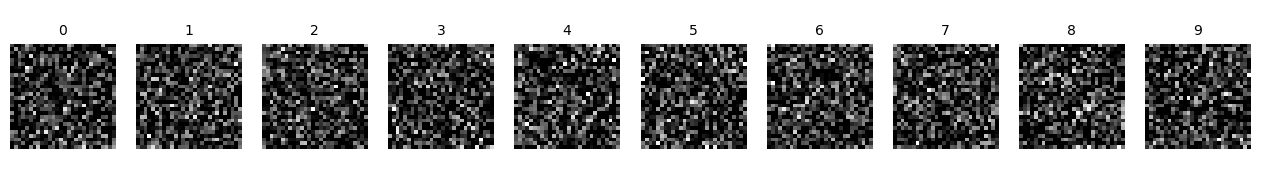}
    \caption{First step of distilled images from 1 out of 10 clients for non-IID federated MNIST with no additions (i.e. soft labels, soft resets, random masking).}
    \label{fig:hard_noniid}
\end{figure}

\begin{figure}[h]
    \centering
    \includegraphics[width=\linewidth]{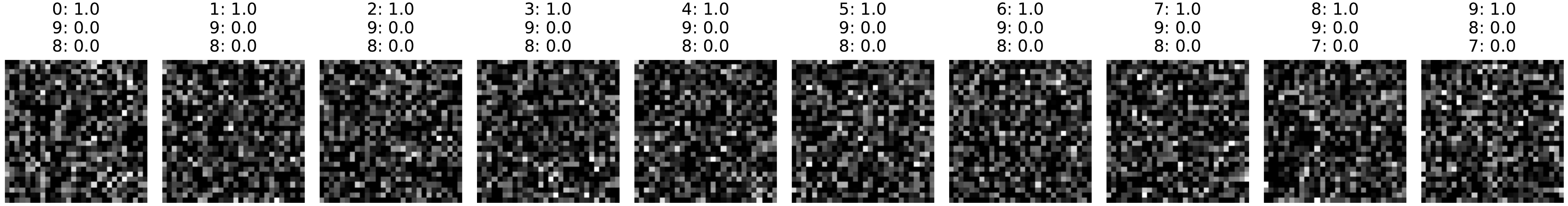}
    \caption{First step of distilled images from 1 out of 100 clients for IID federated MNIST with soft labels. The values above are the 3 labels with the highest logits.}
    \label{fig:soft_iid}
\end{figure}

\subsection{IMDB}\label{app:sents}

We provide a distilled sentence from one of 100 clients for federated IMDB with IID distribution.
The logit is 1.63 for the positive class and 0 for the negative class. 
The corresponding distill learning rate is 0.0272.

\begin{tcolorbox}
shaw malone assembled shelly pendleton tha insanity vietnam finishes morton leather watts respectable mastery funky idle watched peripheral ely glossy 1934 honed periods suppress setting eden arises resides moses aura succumb prc missing dyer angela emulate showcased meredith embraces bonnie translates replicate potts segment affects enhances stein juliet bumping mystic resistance token alienate hays unnamed mira rewarded fateful aspire uniformly bliss mermaid burnt joins unforgettable martino namely marshal ivan morse segment pleads boasting victorian closeness rafael reid saddle boot hawks lingered landon ...
\end{tcolorbox}

Further, we exhibit a distilled sentence for non-IID federated IMDB.
The logit is 1.68 for the positive class and 0 for the negative class. 
The corresponding distill learning rate is 0.0284.
\begin{tcolorbox}

outset wed burroughs grossly contacted reginald anticipating dimitri returns nap housed feeds pitting woodward potts graduates attendant inherit superficial pleasure yanks pills salem tombstone mcintyre finishes ponder pa concede thru herzog getting supports claudio board elevated lieu chaney cashing meantime denise disposition mess whopping comprehend slicing haley cronies screens zombie assures separately ill. debacle helm aroused scrape minuscule dozen wears devoid bio drunken recommendation shrewd denying decaying blocks primal housekeeper moviegoers mates crook useless dictates cap ...
\end{tcolorbox}

\subsection{TREC6}

In addition, we show a distilled sentence from 1 out of 29 clients for federated TREC6 with IID distribution.
The logit is 1.96 for class 1, and 0 for the remaining classes.
The corresponding distill learning rate is 2.25.

\begin{tcolorbox}
conversion loop monster manufactured causing besides stealing yankee 1932 igor supplier nicholas lloyd sees businessman alternate alternate photograph portrayed tale trials 49 principal sequel authors topped donation fictional bull philip
\end{tcolorbox}

At last, we illustrate a distilled sentence from 1 out of 29 clients for federated TREC6 with non-IID distribution.
The logit is 1.58 for class 2, and 0 for the remaining classes.
The corresponding distill learning rate is 1.87.
\begin{tcolorbox}
fair listen programming helps lose remembered block changed classical learning break tap klein stole quick reed solomon mouse extension sisters virtual holmes knight medieval norman newton rider nobel rhode murdered
\end{tcolorbox}

\section{Additional Results}

\subsection{LP-DOSFL}

We provide results for LP-DOSFL on non-federated MNIST tasks in Table~\ref{tab:imdb+trec-serial}: federated IMDB, TREC-6, and Sent140.
Hyperparameters and methodology are identical to those used for regular DOSFL in Section~\ref{sec:exp}, other than the change in distillation order from parallel to serial.

\begin{table*}[t]
    \centering
    \caption{DOSFL and LP-DOSFL performance on federated IMDB, TREC-6, and Sent140. Instead of 10 and 100 clients for \textbf{TREC-6}, we have \textbf{2} and \textbf{29} respectively. Note that, in every case, LP-DOSFL outperforms vanilla DOSFL.}
    \label{tab:imdb+trec-serial}
    \begin{tabular}{l l c c c c}
        \toprule
        Dataset & Setting & \multicolumn{2}{c}{10 clients} & \multicolumn{2}{c}{100 clients} \\
        \cmidrule(l){3-4} \cmidrule(l){5-6}
        & & IID & non-IID & IID & non-IID \\
        \midrule
        IMDB & Vanilla & $81.04\%$ & $79.86\%$  & $71.94\%$ & $70.75\%$\\
        IMDB & LP-DOSFL & $85.07\%$ & $83.28\%$ & $81.65\%$  & $80.78\%$ \\
        \midrule
        TREC-6 & Vanilla & $83.60\%$ & $73.40\%$ & $79.00\%$ & $73.60\%$ \\
        TREC-6 & LP-DOSFL & $86.20\%$ & $74.20\%$  & $82.80\%$ & $75.00\%$ \\
        \midrule
        Sent140 & Vanilla & $78.10\%$ & $78.00\%$ & $73.50\%$ & $69.50\%$ \\
        Sent140 & LP-DOSFL & $81.40\%$ & $80.60\%$  & $77.90\%$ & $74.60\%$ \\
        \bottomrule
    \end{tabular}
\end{table*}

\subsection{Moderate Non-IID}

\begin{figure}[t]
    \centering
    \begin{subfigure}[]{0.49\linewidth}
    \centering
        \includegraphics[width=\linewidth]{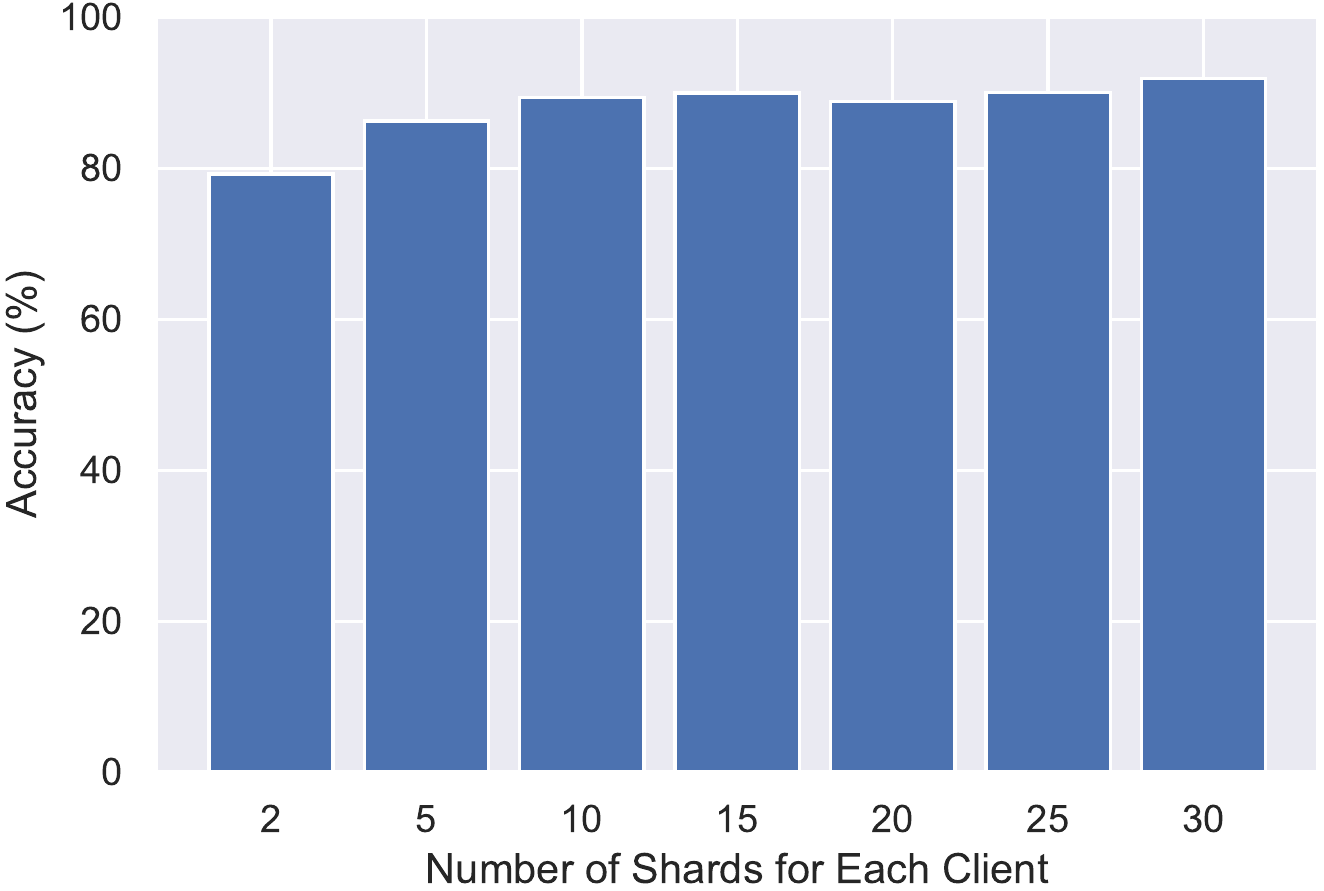}
        \caption{Vanilla DOSFL accuracy for different $s$}
    \end{subfigure}
    \begin{subfigure}[]{0.49\linewidth}
    \centering
        \includegraphics[width=\linewidth]{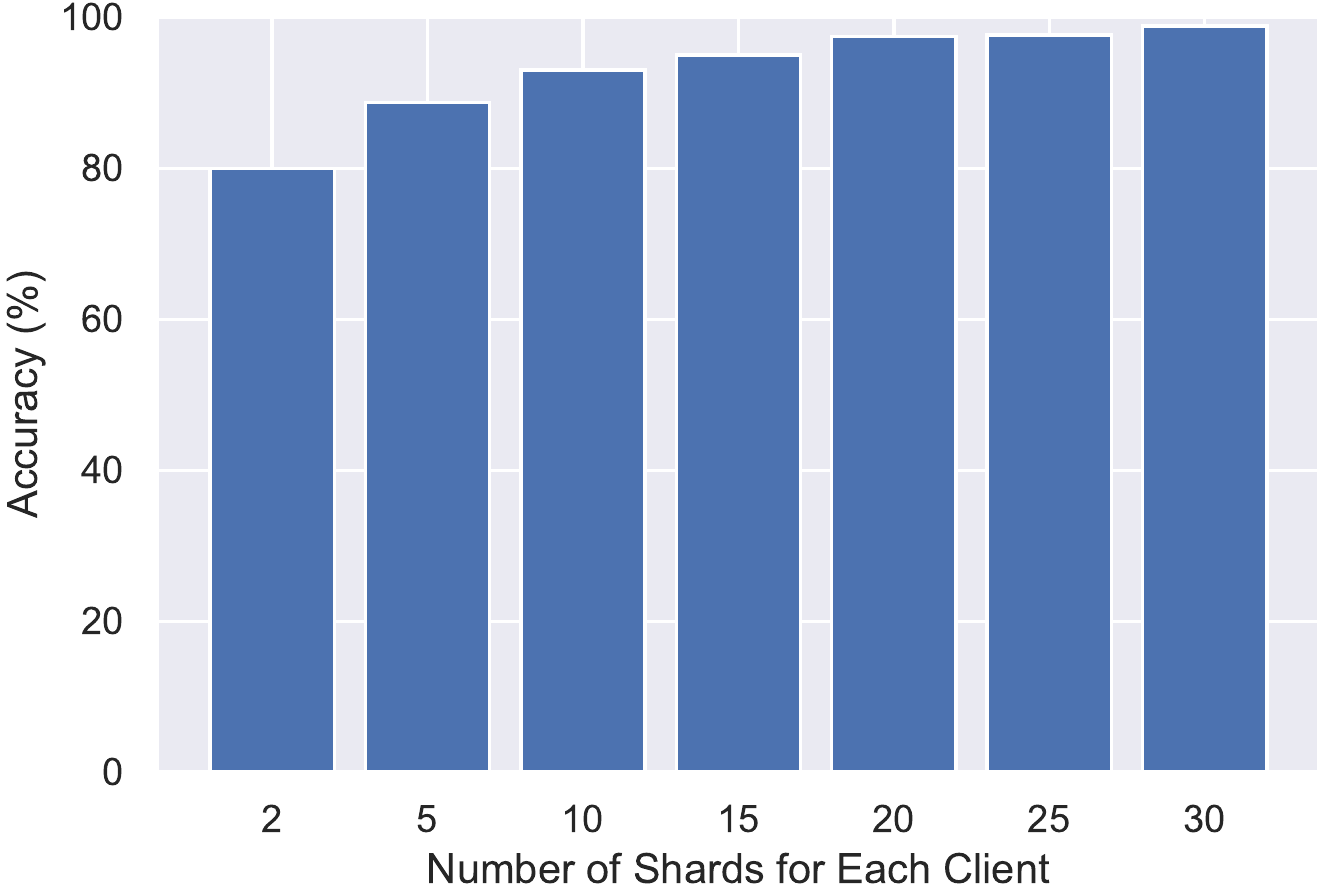}
        \caption{LP-DOSFL accuracy for different $s$}
    \end{subfigure}
    \caption{DOSFL Performance on Federated non-IID MNIST with soft labels and soft resets.}
    \label{fig:acc_shards}
\end{figure}

We perform an analysis of the impact non-IIDness has on DOSFL.
We ran vanilla and LP-DOSFL on 10 client Federated MNIST for shard counts $2$ through $30$.
The results are given in Figure~\ref{fig:acc_shards}.
Importantly, plain DOSFL and LP-DOSFL maintain their IID performance even as the shard count $s$ drops to $10$.
This is a moderately non-IID setting; each client on average still contains examples of all 10 digits.
However, LP-DOSFL slightly curves downward as $s$ decreases while vanilla DOSFL is flat.
Beyond this point, test accuracy degrades quickly until both vanilla and LP-DOSFL have similar test accuracies once $s = 2$.

\end{document}